\documentclass{article}
\usepackage{spconf,amsmath,graphicx, amssymb}
\usepackage{subcaption}
\usepackage[outdir=./]{epstopdf}

\def\M{{\cal M}}
\def\G{{\cal G}}

\title{Denoising-based image reconstruction from pixels located at non-integer positions}
\name{J\'{a}n Koloda, J\"{u}rgen Seiler and Andr\'{e} Kaup\thanks{This work has been partially supported by the Research Training Group 1773 ''Heterogeneous Image Systems'', funded by the German Research Foundation (DFG).}}
\address{Chair of Multimedia Communications and Signal Processing\\
Friedrich-Alexander University Erlangen-N\"{u}rnberg (FAU), Germany}
\begin{document}
%
\maketitle
\begin{abstract}
Digital images are commonly represented as regular 2D arrays, so pixels are organized in form of a matrix addressed by integers. However, there are many image processing operations, such as rotation or motion compensation, that produce pixels at non-integer positions.
Typically, image reconstruction techniques cannot handle samples at non-integer positions. In this paper, we propose to use triangulation-based reconstruction as initial estimate that is
later refined by a novel adaptive denoising framework. Simulations reveal that improvements of up to more than 1.8 dB (in terms of PSNR) are achieved with respect to the initial estimate.
\end{abstract}
\section{Introduction}
\label{sec:intro}

Representing digital images as regular two-dimensional arrays has multiple advantages. First, it allows an efficient implementation of the acquisition hardware. Second, due to its
simple indexing, displaying systems can be implemented more effectively. Moreover, it offers an elegant mathematical tool that is used by various image processing applications.

\begin{figure}[t]
	\centering 
	\begin{subfigure}{0.47\linewidth}
		\includegraphics[width=\textwidth]{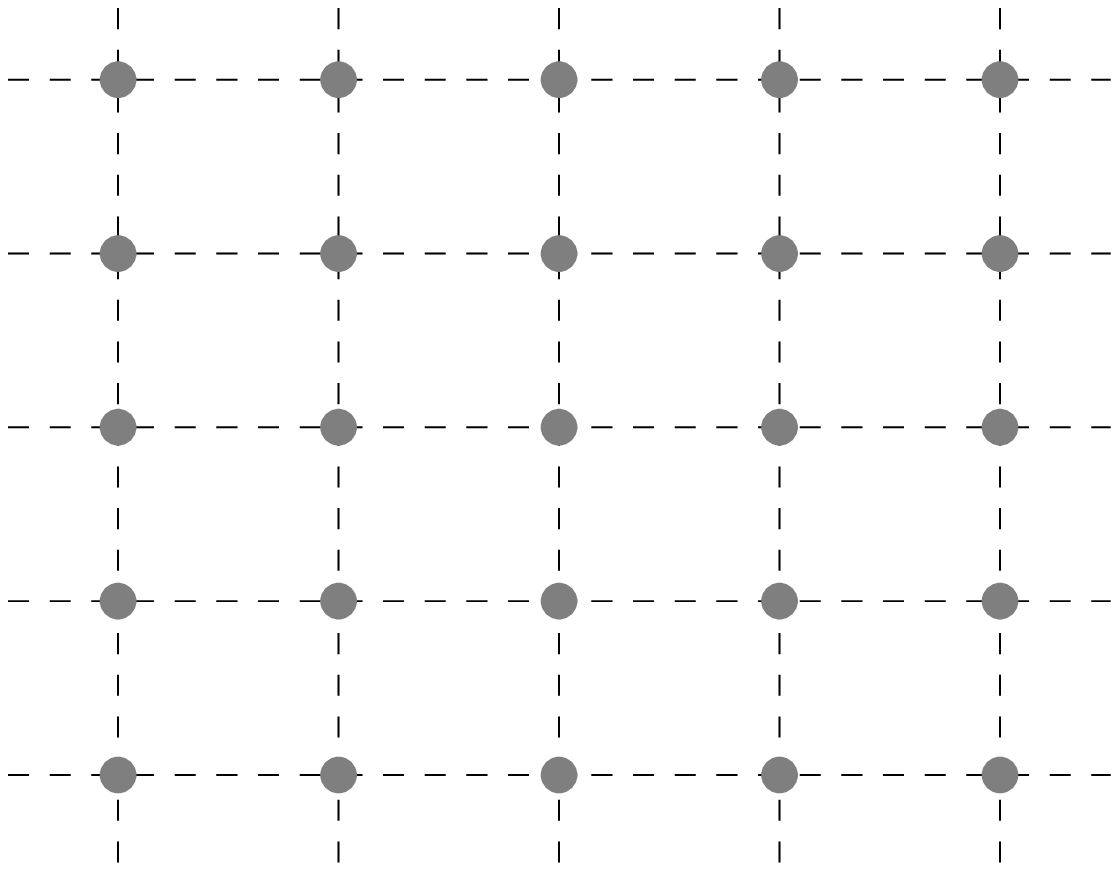}
		\caption{\small{}}
		\label{subfig:regular}
	\end{subfigure}
	\begin{subfigure}{0.47\linewidth}
		\includegraphics[width=\textwidth]{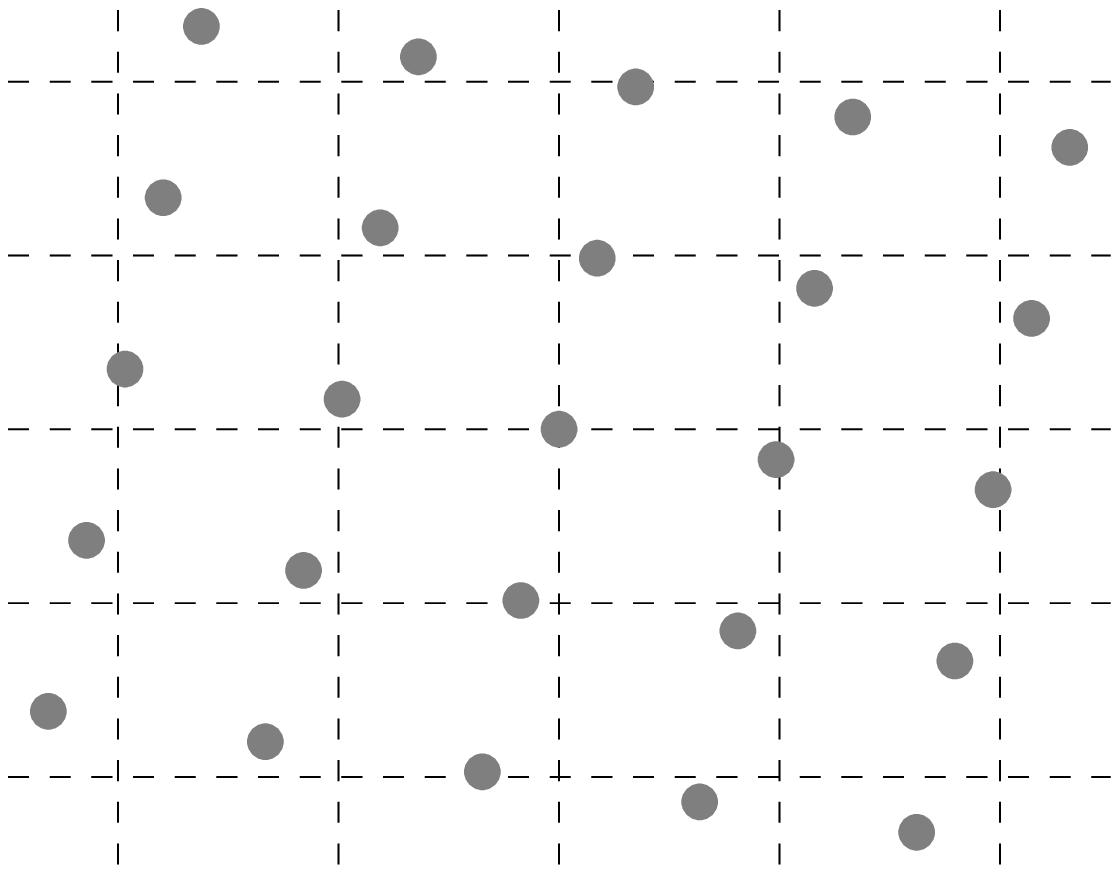}
		\caption{\small{}}
		\label{subfig:floating}
	\end{subfigure}
	\caption{\small{Example of application that yields pixels at non-integer positions. The samples in (b) are obtained by rotating the pixels in (a) by 10 degrees. Dashed lines represent the regular grid.}}
	\label{fig:example}
\end{figure}

However, there are many applications that produce images with pixels located at non-integer positions as their output. Such is the case of super-resolution \cite{Superresolution}, panoramic stitches \cite{Homography},
optical cluster eye \cite{ClusterEye}, image warping \cite{Warping} and motion compensation based on optical flow \cite{Flow}, among others.
Another example of such an application is rotation, as illustrated in Fig. \ref{fig:example}.
Since it is not possible to directly display the rotated samples at non-integer positions, they must be regularized and pixels on the regular
grid have to be estimated. This regularization can be alternatively seen as a reconstruction problem where the unknown pixels on the regular grid are recovered using the available
pixels at non-integer positions.

The vast majority of reconstruction techniques assumes that the input image is already a regular 2D array. This is the case of algorithms based on exemplars \cite{Inpainting, SLPE} or autorregressive modelling \cite{AR}. Reconstruction methods carried out in the transformed domain require the use of regular blocks and cannot be applied either \cite{FOFSE, POCS}. Finally, concealment techniques that exploit local signal statistics, such as gradients and covariances, assume spatial regularity as well \cite{KER}.

In spite of the extensive bibliography on image reconstruction, none of the aforementioned methods is applicable to scenarios where the pixels occupy non-integer positions.
A possible solution comprises the group of techniques based on Delaunay triangulation \cite{Tesselations}. The triangulated space is then used as an input for reconstruction methods that do not require the use of regular grid, such as nearest neighbour method, linear and cubic interpolations \cite{Griddata} and natural neighbour approach \cite{NNI}.
Although triangulation based techniques produce acceptable results, artefacts can be noticed especially for low sample ratios.
A sample ratio is defined as the ratio between the number of available samples and the total amount of pixels an image is comprised of.

In this paper, we propose a novel approach for image reconstruction using samples at non-integer positions. Triangulation-based reconstruction is used as an initial estimate and it is later refined by applying an adaptive denoising procedure. We propose a refinement framework suitable for a wide range of triangulation-based techniques as well as for different types of denoising algorithms.

The paper is organized as follows. Section \ref{sec:den} formally describes the reconstruction problem. In Section \ref{sec:refinement} the generic refinement framework is presented. Experimental results are discussed in Section \ref{sec:results}.
The last section is devoted to conclusions.

\section{Floating mesh reconstruction}
\label{sec:den}

\begin{figure}[t]
	\centering 
	\includegraphics[width=0.4\textwidth]{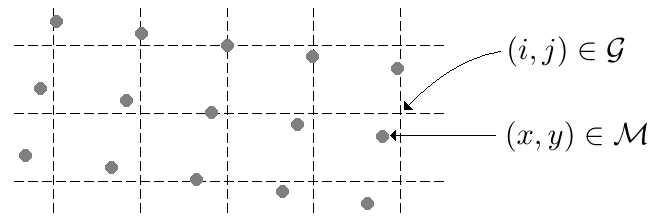}
	\caption{\small{The regular grid $\G$, represented by dashed lines, is addressed by integers $(i,j)$. The floating mesh $\M$ is comprised of samples at non-integer positions $(x,y)$.}}
	\label{fig:definitions}
\end{figure}

Let $\G$ denote the regular grid that represents the desired image and let $\M$ be the floating mesh that contains available samples at non-integer positions, as shown in Fig. \ref{fig:definitions}.
The unknown pixels in $\G$ are reconstructed using the information provided by the available samples in $\M$.
Without loss of generality, we will assume that the image on the regular grid is estimated by any of the aforementioned triangulation based techniques.
Although these techniques can yield an acceptable image quality, a certain amount of reconstruction error is usually present.
Therefore, the estimated image $\tilde{I}$ can be regarded as a noisy version of the original image $I$, i.e.
\begin{equation}
	\tilde{I}(i,j) = I(i,j) + \eta(i,j) \quad \forall (i,j) \in \G .
	\label{noisy_pixel}
\end{equation}
This alternative point of view suggests that a denoising procedure can be applied in order to enhance the quality of the reconstructed image. Note that cancelling the noise $\eta$ is equivalent to reducing the reconstruction error. 

Denoising techniques usually assume that the noise has known and constant statistics \cite{BM3D, TV}. Typically, an image is assumed to be corrupted by a Gaussian noise with known mean
and variance. Having an estimate of the noise statistics permits to set the strength of the denoising procedure \cite{BM3D}. Generally, the higher the noise power, the stronger the denoising applied.

In our reconstruction framework the noise is locally highly non-stationary and, in general, cannot be considered Gaussian.
It depends on various factors such as the reconstruction technique applied or the sample ratio.
The most advanced denoising algorithms cannot handle non-Gaussian noise satisfactorily \cite{NonGaussianNoise} so modifications have to be made.
If we could locally
estimate the noise power (i.e. the reconstruction error) we could adapt the denoising strength so that high errors would be corrected by strong denoising while small errors would be corrected only slightly or not at all.

In the next Section we propose a new approach to estimate the reconstruction error for triangulation based techniques.
This estimation is then employed to control the denoising strength. Although our proposal is applicable to a wide range of denoising techniques we will consider the block-matching and 3D filtering (BM3D) algorithm \cite{BM3D} throughout the paper since it is one of the most efficient denoising methods \cite{NonGaussianNoise}.

\section{Denoising-based refinement}
\label{sec:refinement}

Assuming a zero-mean generic noise, its power $\sigma^2$ is locally defined as the expectation of the squared reconstruction error $\varepsilon$, i.e.

\begin{equation}
  \begin{array}{l l}
	\sigma^2(i,j) = &  E\{\varepsilon(i,j)\} = E \bigg\{ \left( \tilde{I}(i,j) - I(i,j) \right)^2 \bigg\} \\
	& \forall (i,j) \in \G
  \end{array} .
  \label{error}	
\end{equation}
From this equation emanate the following issues:
\vspace{-0.15cm}
\begin{enumerate}
	\item The error cannot be computed directly since the original image $I$ is unknown.
	\vspace{-0.15cm}
	\item BM3D (and basically any denoising algorithm) is originally not designed for reconstruction purposes. It assumes stationary Gaussian noise which, as discussed in the previous section,
	is not true in general in our reconstruction framework. Therefore, the value of noise power $\sigma^2$, that controls the denoising strength, cannot be used directly as input for BM3D. Instead, this value must be carefully selected in order to maximize the reconstruction quality.
\end{enumerate}
\vspace{-0.15cm}

The error estimation and the selection of the optimal value of $\sigma^2$ are the two issues that are addressed in the next subsections.

\subsection{Estimation of the reconstruction error}
\label{subsec:estimation}

In order to estimate the reconstruction error, we will consider the following two assumptions:
\vspace{-0.15cm}
\begin{enumerate}
	\item The higher the density of available samples the lower the reconstruction error \cite{AR}. In other words, high quality reconstructions are achieved for pixels
	 surrounded by large amount of available samples on the floating mesh.	
	 \vspace{-0.15cm}
	\item Given the high spatial correlation of natural images, samples close to the missing pixel are more relevant for estimation than distant samples \cite{FOFSE}.
\end{enumerate}
\vspace{-0.15cm}
In order to take into account both criteria, the amount of effective data $\xi$ around the missing pixel is considered. We propose the following function to compute this amount,

\begin{equation}
	\xi(i,j) = \sum_{(x,y) \in \M} e^{-\sqrt{(i-x)^2 + (j-y)^2}} \quad \forall (i,j) \in \G.
	\label{effective_data}
\end{equation}
Note that unlike the reconstruction error $\varepsilon$, the amount of effective data can be computed without any knowledge about the reference image.

\begin{figure}[t]
	\centering 
	\includegraphics[width=0.5\textwidth]{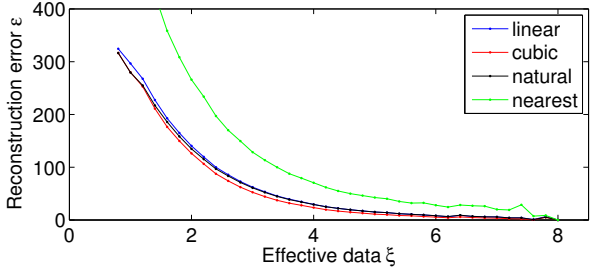}
	\caption{\small{Average reconstruction error $\varepsilon$ for different amounts of effective data $\xi$ using triangulation with linear, cubic and natural neighbour interpolations as well as the nearest neighbour approach.}}
	\label{fig:errors}
\end{figure}

In order to analyze the suitability of using $\xi$ as an indicator of the true reconstruction error $\varepsilon$, this error is shown in {Fig. \ref{fig:errors}} as a function of the
amount of effective data $\xi$. For this, Tecnick image database \cite{Tecnick} has been employed.
Note that since these images are used here for validation purposes, they will not be used later for testing the performance of our proposal.
As shown in Fig. \ref{fig:errors}, the higher the amount of effective data the lower the reconstruction error.
It follows that the amount of effective data around the pixel that is being reconstructed is a reliable indicator of the reconstruction error.
The amount of effective data can be used to select the optimal denoising strength in order to maximize the reconstruction quality. This selection is detailed in the next subsection.

\subsection{Denoising strength selection}
\label{subsec:strength_selection}

\begin{figure}[t]
	\centering 	
	\includegraphics[width=0.5\textwidth]{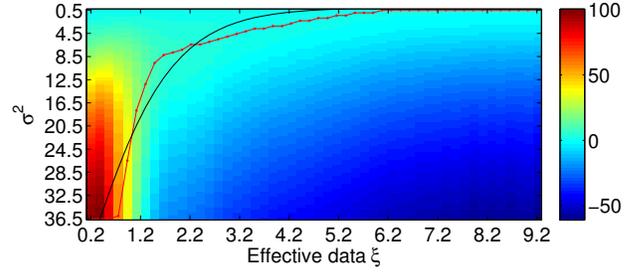}
	\caption{\small{Average gain per pixel for different values of $\xi$ and $\sigma^2$. The path of maximum gain is depicted by red colour. Its sigmoid approximation is shown in black.}}
	\label{fig:gain}
\end{figure}

We still have to find the link between the optimal denoising strength, $\sigma^2_{opt}$, and the amount of the effective data $\xi$. In order to do so, we define the gain, $G$,
of the denoised image with respect to the initial estimate,

\begin{equation}
	\begin{array}{l l l}
		G & = & \varepsilon(i,j) - \varepsilon_D(i,j) \\
		  & = & \left( I(i,j) - \tilde{I}(i,j) \right)^2 - \left( I(i,j) - \tilde{I}_{D(\sigma^2)}(i,j) \right)^2 \\
		 & & \forall (i,j) \in \G.
	\end{array}
\label{gain}
\end{equation}
where $\tilde{I}_{D(\sigma^2)}$ represents the initial estimate $\tilde{I}$ denoised by BM3D using $\sigma^2$ as the noise power parameter.
Figure \ref{fig:gain} shows the average gain per pixel as a function of $\xi$ for different values of $\sigma^2$.
Cubic interpolation is used as the initial estimate.
In order to estimate the relation $\sigma^2_{opt} = f(\xi)$, we propose to maximize the gain. Thus, given an amount of effective data, the corresponding $\sigma^2_{opt}$ that maximizes the gain is selected.
The path of maximum gain is also shown in Fig. \ref{fig:gain}. It suggests an inverted sigmoid-like relation where $\sigma^2_{opt}$ saturates as $\xi$ increases. In other words, if there is enough effective data around the missing pixel only a slight denoising is required. On the other hand, low amounts of effective data require strong denoising to improve the reconstruction quality. Thus, in order to link $\sigma^2_{opt}$ with the amount of effective data, we propose the following sigmoid shaped function,

\begin{equation}
\sigma^2_{opt}(i,j) = \alpha \left( 1 - \frac{1}{1 + e^{-\left( \frac{\xi(i,j)}{\gamma} + \beta \right) }} \right)
\label{regression}
\end{equation}
where $\alpha$ indicates the scale, $\beta$ sets the offset and $\gamma$ controls the decay.
These parameters are estimated by non-linear regression using the Tecnick database. The computed values are detailed in Section \ref{sec:results}.

Finally, the reconstruction algorithm is summarized as follows:
\vspace{-0.15cm}
\begin{enumerate}
	\item Obtain an initial image estimate by applying a triangulation based technique.
	\vspace{-0.15cm}
	\item For every pixel on the regular grid, compute the amount of effective data $\xi$ of Eq.(\ref{effective_data}).
	\vspace{-0.15cm}
	\item Use $\xi$ to estimate the optimal denoising strength $\sigma^2_{opt}$ according to Eq.(\ref{regression}).
	\vspace{-0.15cm}
	\item Finally, for every pixel in $\G$ apply BM3D with the noise power parameter set to the corresponding value of $\sigma^2_{opt}$.
\end{enumerate}

\section{Simulation Results}
\label{sec:results}

In order to test the performance, both the original image and the floating mesh need to be known. Since, to the best of our knowledge, there is no such database available, we design the following framework.
The ARRI image dataset \cite{ARRI} of 11 images with dimensions of 2880$\times$1620 is employed.
The images are downsampled by taking one of every
$\phi$ pixels in each dimension, where $\phi$ is the downsampling factor. We use $\phi = 5$ in our simulations. Note that previous to downsampling a low-pass antialiasing filter with
the cut-off digital frequency of 1/$\phi$ is applied. We will then use the downsampled images as references.
Thus, between two consecutive pixels in a downsampled image, there are $(\phi - 1)$ intermediate samples that can be obtained from the original image (previously filtered with the
antialiasing filter). These intermediate pixels comprise the floating mesh. The pixels from the floating mesh are randomly sampled and they are used for reconstructing the
pixels on the regular grid, i.e. for reconstructing the reference image.

\begin{figure}[t]
	\centering 	
	\includegraphics[width=0.5\textwidth]{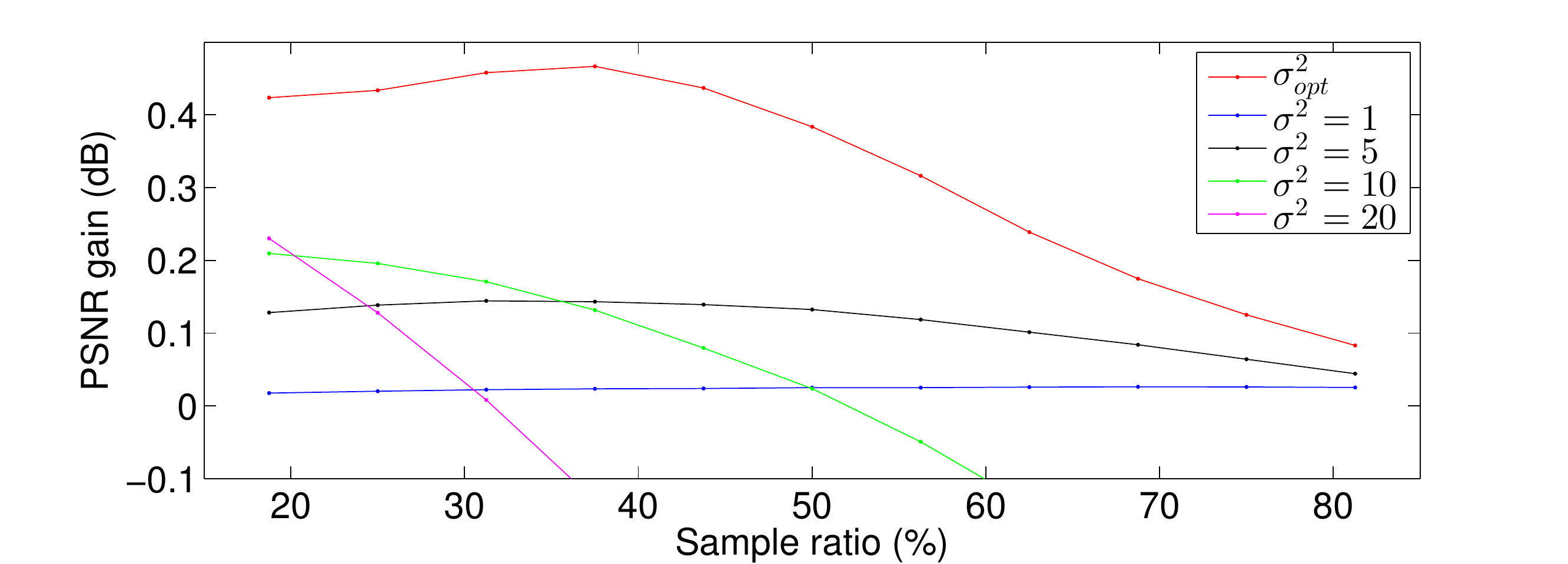}
	\caption{\small{Average PSNR gain (in dB) for different sample ratios using BM3D denoising algorithm with fixed $\sigma^2$ value and the proposed adaptive approach with $\sigma^2_{opt}$. Cubic interpolation is used as initial estimate.}}
	\label{fig:gain_comparison}
\end{figure}

Sample ratios from 20\% up to 80\% are used.
The initial estimates are computed using triangulation based linear interpolation (LI) \cite{Griddata}, cubic interpolation (CI) \cite{Griddata},
natural neighbour interpolation (NI) \cite{NNI} and nearest neighbour approach (NN) \cite{Griddata}.
For every pixel the corresponding value of $\sigma^2_{opt}$ is computed according to Eq.(\ref{regression}). As already mentioned,
the parameters $\alpha$, $\beta$ and $ \gamma$ are estimated by non-linear regression using the Tecnick database and are summarized in Table \ref{tab:parameters}.
Note that the Tecnick database is independent from the ARRI dataset which is used to test the performance.

\begin{table}
	\begin{tabular*}{0.48\textwidth}{l| @{\extracolsep{\fill}} cccc}	
		\hline\hline
		& LI             & CI             & NI            & NN              \\ \hline
		$\alpha$         & \small{73.69}  & \small{105.54} & \small{44.14} & \small{84.38}   \\		
		$\beta$          & \small{-0.71}  & \small{0.08}   & \small{-3.64} & \small{0.40}    \\		
		$\gamma$         & \small{0.68}   & \small{0.97}   & \small{0.27}  & \small{4.17}    \\		
		\hline\hline	
	\end{tabular*}
	\caption{\small{Estimated parameters of $\alpha$, $\beta$ and $\gamma$ for BM3D denoising algorithm.}}
	\label{tab:parameters}
\end{table}

\begin{table}
	\begin{tabular}{l|cccccccc}		
		\hline\hline
		& \multicolumn{7}{c}{Sample ratio} \\
		& 20\%          & 30\%          & 40\%          & 50\%          & 60\%          & 70\%          & 80\%             \\ \hline
		LI & \small{0.25}  & \small{0.32}  & \small{0.30}  & \small{0.20}  & \small{0.11}  & \small{0.05}  & \small{0.02}  \\
		CI & \small{0.43}  & \small{0.46}  & \small{0.46}  & \small{0.38}  & \small{0.27}  & \small{0.17}  & \small{0.09}  \\
		NI & \small{0.14}  & \small{0.17}  & \small{0.09}  & \small{0.03}  & \small{0.01}  & \small{0.01}  & \small{0.00}  \\
		NN & \small{1.82}  & \small{1.78}  & \small{1.72}  & \small{1.66}  & \small{1.58}  & \small{1.50}  & \small{1.42}  \\ \hline\hline
	\end{tabular}
	\caption{\small{Average PSNR gain (in dB) for different sample ratios using BM3D denoising algorithm.}}
	\label{tab:results}
\end{table}

Since the proposed technique can be considered as a generic posterior refinement procedure over an initial estimate, it makes sense to evaluate the performance in terms of gain.
Let us first analyze the advantage of applying the proposed adaptive denoising over the use of a fixed $\sigma^2$ value.
Figure \ref{fig:gain_comparison} shows that strong denoising (large values of $\sigma^2$) improves the performance for low sample ratios but deteriorates the image quality for higher sample ratios where the initial estimate is already expected to yield a relatively small reconstruction error.
This issue is avoided by employing mild denoising (small values of $\sigma^2$) but only a minor improvement is achieved.
As shown in Fig. \ref{fig:gain_comparison}, the proposed adaptive procedure outperforms the fixed-valued denoising in terms of PSNR and provides considerably better results over the entire sample ratio range.

\begin{figure}[t]
	\centering 
	\begin{subfigure}{0.32\linewidth}
		\includegraphics[width=\textwidth]{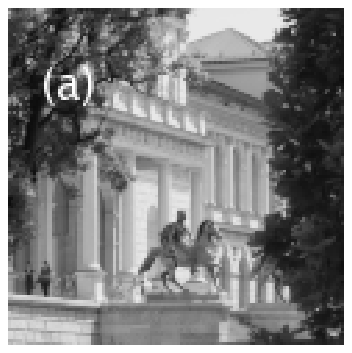}
	\end{subfigure}
	\begin{subfigure}{0.32\linewidth}
		\includegraphics[width=\textwidth]{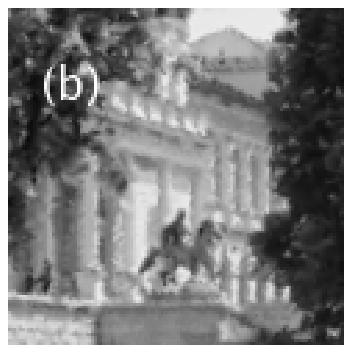}
	\end{subfigure}
	\begin{subfigure}{0.32\linewidth}
		\includegraphics[width=\textwidth]{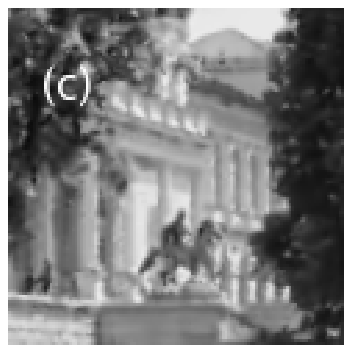}
	\end{subfigure}
	\caption{\small{Subjective comparison of the reconstruction quality using a sample ratio of 50\%. (a) Original image. (b) Reconstructed by CI (PSNR = 27.22 dB). (c) Reconstructed by the proposed method using BM3D (PSNR = 27.43 dB).}}
	\label{fig:akademie}
\end{figure}

\begin{figure}[t]
	\centering 
	\begin{subfigure}{0.32\linewidth}
		\includegraphics[width=\textwidth]{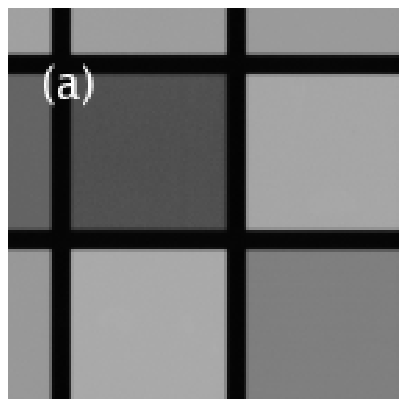}
	\end{subfigure}
	\begin{subfigure}{0.32\linewidth}
		\includegraphics[width=\textwidth]{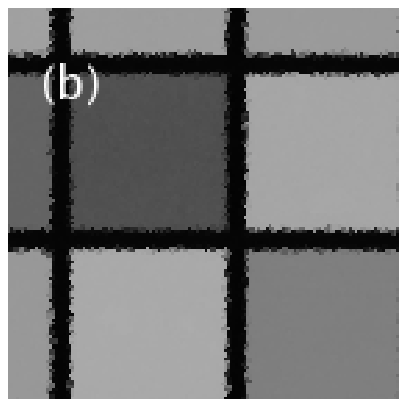}
	\end{subfigure}
	\begin{subfigure}{0.32\linewidth}
		\includegraphics[width=\textwidth]{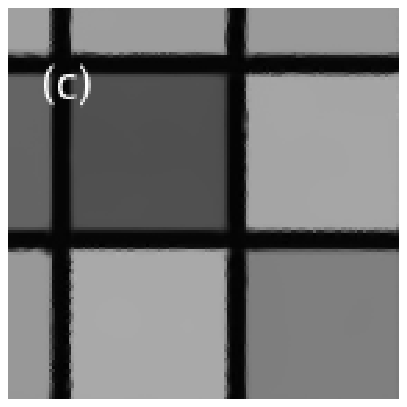}
	\end{subfigure}
	\caption{\small{Subjective comparison of the reconstruction quality using a sample ratio of 50\%. (a) Original image. (b) Reconstructed by NN (PSNR = 26.56 dB). (c) Reconstructed by the proposed method using BM3D (PSNR = 28.57 dB).}}
	\label{fig:sharpness}
\end{figure}

Table \ref{tab:results} shows the average gain of the proposed framework with respect to LI, CI, NI and NN.
It follows that for high sample ratios the improvement is relatively moderate given that the initial estimate is already of high quality.
On the other hand, for low sample ratios the image quality is considerably improved.
The proposed framework can improve the average reconstruction quality up to 1.82 dB depending on the initial estimate. This improvement is also seen at subjective level, as illustrated in Fig. \ref{fig:akademie} and Fig. \ref{fig:sharpness}.

\section{Conclusions}
\label{sec:conclusions}

In this paper, a scenario of pixels located at non-integer positions is considered. These samples are used to estimate the pixels on the regular grid.
We have proposed a novel framework for applying denoising techniques in order to improve the reconstruction quality.
An initial estimation is obtained by means of triangulation and is refined by applying an adaptive denoising approach. The proposed method adapts the denoising strength according to the amount of effective data around the missing pixel. Simulations reveal that improvements of more than 1.8 dB can be achieved. 

Ongoing work is focused on extending the proposed framework to other denoising algorithms and on testing it on various image processing applications.

\bibliographystyle{IEEEbib}
\bibliography{refs}

\end{document}